# 3D Facial Action Units Recognition for Emotional Expression

Norhaida Hussain[1], Hamimah Ujir, Irwandi Hipiny and Jacey-Lynn Minoi[2]

[1]Department of Information Technology and Communication, Politeknik Kuching, Sarawak, Malaysia
[2]Faculty of Computer Science and Information Technology, Universiti Malaysia Sarawak, Kota Samarahan, Sarawak, Malaysia

The muscular activities caused the activation of certain AUs for every facial expression at the certain duration of time throughout the facial expression. This paper presents the methods to recognise facial Action Unit (AU) using facial distance of the facial features which activates the muscles. The seven facial action units involved are AU1, AU4, AU6, AU12, AU15, AU17 and AU25 that characterises happy and sad expression. The recognition is performed on each AU according to rules defined based on the distance of each facial points. The facial distances chosen are extracted from twelve facial features. Then the facial distances are trained using Support Vector Machine (SVM) and Neural Network (NN). Classification result using SVM is presented with several different SVM kernels while result using NN is presented for each training, validation and testing phase.

**Keywords:** Facial action units recognition, 3D AU recognition, facial expression

## 1. INTRODUCTION

Facial expressions are significant to non-verbal communications as they provide varies information such as emotional state, attitudes, personality and age. The analysis of facial expression plays a major role in emotional recognition that contributes to the development of Human-Computer Interaction (HCI) system which could be applied to area such as neurology, lie detection, intelligent environments and paralinguistic communication [1]. There are six basic emotional facial expressions acknowledged by psychologists; happy, surprise, angry, fear, sad and disgust. However, the six emotions partially represent human facial expressions.

Facial expression is initiated by the contractions of distinct facial muscles that caused temporary deformation of facial surface including the geometry and texture characteristics [2]. Action Unit (AU) defined in the Facial Action Coding System (FACS) is a component of facial expression triggered by different facial muscle that lies below facial skin.

This paper presents the study of the AU presence in corresponds to happy and sad expression using two different classification approaches. The first objective is to compute 3D facial distances for happy and sad expression. The second objective is to classify the AU using Support Vector Machines (SVM) and Neural Networks (NN). Section 2 describes the related works in this field, followed by a discussion on the AU recognition process in section 3. Then, the geometric facial features are discussed in section 4. Section 5 is about the experiments.

## 2. RELATED WORKS

The presence of AU6, AU12 and AU25 indicates happy facial expression and those AUs are presence only at certain duration throughout the full facial expression, which is from neutral to offset phase. To improve recognition performance of each AU, selecting salienfacial feature points is very important since they represent the geometric property of AU activation. 3D facial distances is used [15] to measure facial expression intensity. In [12][14], surface normal are computed using the provided 3D facial points to classify facial expression.

On the other hand, the classifier algorithms used to classify the AUs also plays an important rate in getting good results. Thus, it is important to decide on which classifier that suitable to recognise the AUs. There are many techniques used to classify AUs from 3D facial such as SVM [3], Hidden Markov Model (HMM) [4] and Local Binary Pattern (LBP) [5] that achieved good classification result.

TABLE I. EMOTIONS AND CORRESPONDING AUs.

| Action Units | Ekman & Friesan[6] | Lucey et al [7] | Karthick et al [8] |
|---|---|---|---|
| Happy | 6+12 | 6+12+25 | 6+12+25 |
| Sad | 1+4+15 | 1+2+4+15+17 | 1+4+15+17 |
| Surprise | 1+2+5B+26 | 1+2+5+25+27 | 5+26+27+1+2 |
| Disgust | 9+15+16 | 1+4+15+17 | 9+17 |
| Anger | 4+5+7+23 | 4+5+15+17 | 4+5+7+23+24 |
| Fear | 1+2+4+5+20+26 | 1+4+7+20 | 4+1+5+7 |

All basic facial expressions involve different facial muscles. The AUs correspond to all basic facial expressions presented in Table I are based on previous studies by [6]-[8]. These AUs are activated from predefined facial features and normally the distance between two or more facial points are the facial features used. These distances are the features that are extracted to recognise the activation of AU during any facial expression. Based on Table I, happy and sad expression involved different set of facial muscles. Table II defines AU which involved in Happy and Sad expression.

TABLE II. AUs DEFINITION [9].

| AU | Definition |
|---|---|
| AU1 | Raised inner eyebrow |
| AU4 | Eyebrows drawn together, lowered eyebrows |
| AU6 | Raised cheek, compressed eyelid |
| AU12 | Mouth corners pulled up |
| AU15 | Mouth corners downward |
| AU17 | Chin raised |
| AU25 | Lips parted |

The AU recognition rule could affect the accuracy rate of AU presence during occurrence of any facial expression. In the previous work by [9], 30 FACS rules and 30 expert system rules are mapped to recognise the activation of a single AU. Each AU is encoded with certain rule based on the distance of facial points, measured between expression and expressionless face. The encode rules which related to the AUs correspond to happy and sad facial expression are described in [9] and they are revised to ensure it is suitable with the selected facial points used in this project. For example, by referring to a face model used in [9], an increased in vector of BAD and B1A1D1 indicates the activation of AU1.

A total of 27 AUs was successfully detected with average recognition rate of 86.6% using 24 rules developed for profile-view face of face image sequences [10]. AU recognition rules was developed based on 15 fiducial points on profile-view face image. In addition, the rule to detect the temporal segments of AUs is also included in AU recognition rule. Based on Table 2, Rule 1 is defined to detect the presence of AU1 and this AU is only considered as presence when position of a facial feature located at inner corner of eyebrow, is moving up or down more than the threshold value which is equal to 1.

TABLE III. RULES TO RECOGNISE THE PRESENCE OF AU1, AU4, AU6, AU12, AU15, AU17 AND AU25 [10].

| Rule | Description |
|---|---|
| Rule 1 | IF up/down (P12) > ε THEN AU1 - p |
| Rule 3 | IF [inc/dec(P2P12)] > ε AND inc/dec(P2P12) ≤ T1 THEN AU4 - p |
| Rule 5 | IF [inc/dec(P13P14)]$_t$ > ε AND inc/dec(P13P14) ≤ T1 AND up/down(P7) > ε THEN AU6 - p |
| Rule 9 | IF up/down(P7) > ε AND inc/dec(P5P7)] $_<$ ε THEN AU12-p |
| Rule 11 | IF up/down(P7) < ε THEN AU15-p |
| Rule 13 | IF NOT (AU28 OR AU 28t OR AU28b) AND inc/dec(P10P15) > ε THEN AU17- |
| Rule 18 | IF [inc/dec(P6P8)]$_t$ < ε AND inc/dec(P4P10) ≥ ε THEN AU25 - p |

## 3. AU RECOGNITION PROCESS

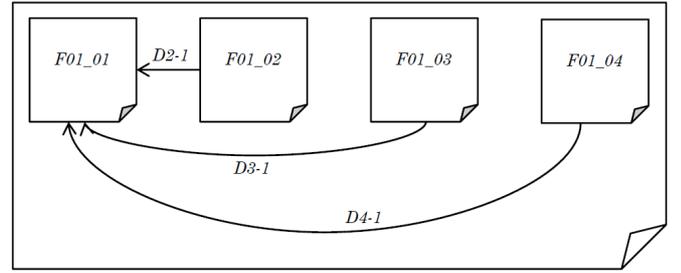

Fig. 2. Illustration of displacement facial points' measurement

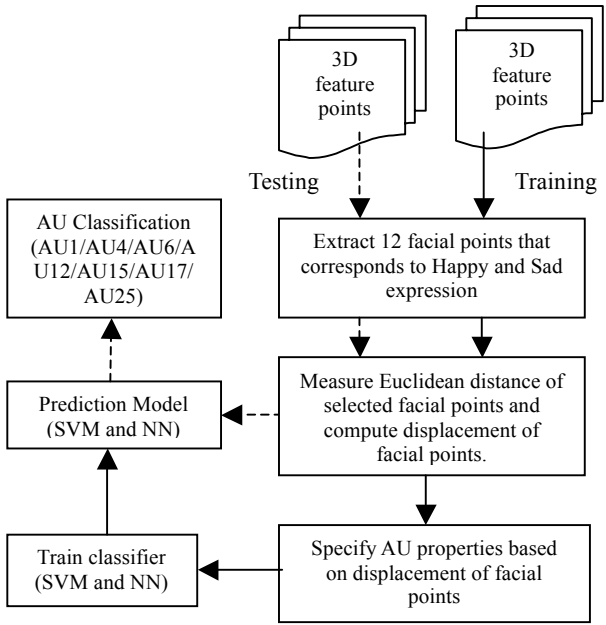

Fig. 1. The block diagram of process

Concentrating on a region of interest (ROI) will further help to recognize an AU. For example, to measure human task performance, ROI of gaze as is used [16][17]. Our ROI in this study is the region of the facial points that involve in face muscle deformation. There are seven main processes involved in this project, as described in figure 1. The next process after facial point's extraction is the distance measurement between two facial points.

The 3D feature points are extracted from BU-4DFE database to obtain twelve facial points that corresponds to happy and sad expression. For each AU, the properties are specified based on increasing or decreasing of distance value between selected facial points.

The next process is to train the learning classifier to obtain the prediction model. There are many classification algorithms that are used in facial expression and AU recognition such as boosting, SVM, NN, nearest neighbour, Naives Bayes and random forests.

The input to the learning classifier is the distance value for each facial point. To obtain the displacement value for each facial point, the distance is computed between the 30 second, third and fourth intensity level to the first intensity level. Based on figure 2, facial data labelled as F01_01, F01_02, F01_03 and F01_04 are the intensity level consisted in the first facial data. The distances for all twelve properties are computed before the displacement of facial points between frame F01_02 to F01_01, FO1_03 to F01_01 and F01_04 to F01_01.

As shown in figure 2, the displacement of facial points in the second intensity level (F01_02) is computed and the distances are saved as D2-1. Then the displacement of facial points in the third intensity level (F01_03) is computed and the results are saved as D3-1. Lastly, the displacement of facial points in the fourth intensity level (F01_04) is computed and the distances are saved as D4-1. The distances are measured for all facial data which consists of 30 facial data for happy and 30 facial data for sad expression.

The next process is to identify the properties for each AU according to displacement of facial points measured earlier. The distance and corresponding AU label are the input to the learning classifier during training. The AU is labelled either as 1 or 0 that describes the presence and not presence of AU respectively. For AU1, AU6 and AU25, the input to the learning classifier is the displacement distance and the expected output is the AU labelled as 1 or 0. Meanwhile for AU4, AU12 and AU17, there are two distance values with one expected output. The expected output which is the AU label is obtained by using logical AND operator for both properties. If property 1 and property 2 is equal to 1, AU is labelled as 1 which indicates AU is presence. If either one of the property equal to 1, AU label is set to 0. Both properties must be satisfied for the AU to be presented. Lastly, AU15 has three distance properties as input to the classifiers.

NN is trained using back-propagation algorithm and the training process stops when generalization stops improving. However, if the percent error is high, the network is trained again to achieve better results with less missed classification. Three different kernels are used in SVM classification.

## 4. GEOMETRIC FACIAL FEATURES

The facial expression analysis in 3D facial could be performed by extracting different features such as geometric which includes facial points and position of the facial points. Happy and sad facial expressions involve only several AUs, which means happy and sad expressions only involve certain facial points. There are twelve facial points extracted from the original 83 facial points provided in BU-4DFE database. Geometric features are computed between the selected facial points.

The idea of selecting facial points only one side of the face is based on [11]. In this study, one extra point which is point 76 that represents tip of the chin is added. Tip of the chin is used to determine the property of AU17 (chin raised) which occurs during sad expression. The position of selected facial points is observed for each frame. The distances of selected points are described in Table 3.

TABLE IV. EMOTIONS AND CORRESPONDING AUs.

| Facial Expression | AU | Distance Properties |
|---|---|---|
| **Happy** | AU6 | - P11P15 (d-) |
| | AU12 | - P55P82 (d-) |
| | | - P45P55 (d-) |
| | AU25 | - P52P58 (d+) |
| **Sad** | AU1 | - P9P27 (d+) |
| | AU4 | - P9P27 (d-) |
| | | - P27P28 (d-) |
| | AU15 | - P45P55 (d+) |
| | | - P55P82 (d+) |
| | | - P9P55 (d+) |
| | AU17 | - P76P82 (d+) |
| | | - P76P45 (d+) |

The Euclidean distance is computed between the selected facial points for each frame in happy and sad facial expression. The similar distance calculation is found in [12][13][15]. The formula to compute the Euclidean distance, ED, of point label x and y is:

$$ED(x, y) = \sqrt{(x_1 - y_1)^2 + (x_2 - y_2)^2 + (x_3 - y_3)^2} \quad [1]$$

The difference of distance for each distance property is observed from the second frame until the last frame. Difference of distance between those frames is compared to the first frame. The increased (d+) and decreased (d-) of those distance is observed to recognise AU as shown is Table 3.

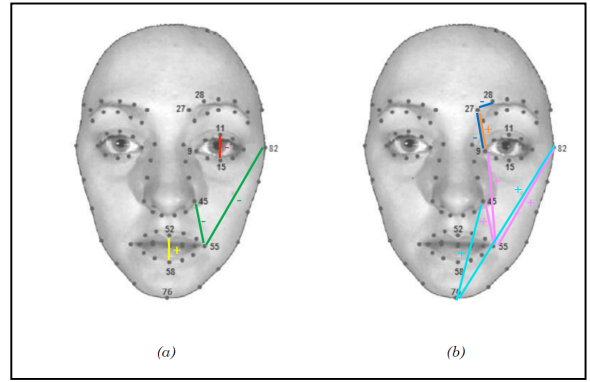

Fig. 3. The distance properties (a) Distance properties for Happy expression (b) Distance properties for Sad expression.

The presence of AU is set based on distance properties shown in Table 3. For example, AU12 (lip corner pulled up) is activated when both distance properties are TRUE and the value is set to 1. If both properties are not achieved, the value is set to 0 which means AU is not activated. AU12 is activated when there is decreased distance of P55P82 and P45P55. P55 is the point at left corner of mouth contour and P82 is the point of left face contour close to left eye contour. If the facial expression is happy, AU12 is activated when the lip corner pulled up results in decreasing distance between lip corner and face contour on the same side of the face. The distance properties for each AU are observed for each frame. The distance difference and AU presence value (0 or 1) are the input for classifier. The value of AU either 1 or 0 is the expected output for the learning process.

## 5. EXPERIMENTS

The BU-4DFE data consists of four intensity levels for each frame. To obtain the difference of position for each facial point, the distance is calculated between the second, third and fourth intensity level frame to the first intensity level frame.

The experiments are carried out using 5-fold. Table 4 presents the result of the experiments. Average correct classification for all AUs is 99.91%, 99.91% and 99.89% is obtained using Linear, Gaussian and Quadratic kernel respectively. For AU1, 100% is achieved using Quadratic kernel while Linear and Gaussian achieved 99.99% and 99.98% respectively. Each AU is correctly classified by the rate more than 90% for the three kernels. classification rate achieved using Quadratic kernel is 100% (AU1), 90.22% (AU12), 86.67% (AU15) and 92% (AU17). While for AU6, AU15 and AU25 the best kernel to use is Linear with 96.89%, 86.67% and 94.67% respectively. Lastly, AU4 is best classified using Gaussian kernel with 94.67% which is 0.45% greater than Quadratic kernel.

TABLE V. AVERAGE CORRECT CLASSIFICATION USING SVM

| AU | Linear | Gaussian | Quadratic |
|---|---|---|---|
| AU1 | 98.67 | 98.22 | 100 |
| AU4 | 85.78 | 94.67 | 94.22 |
| AU6 | 96.89 | 93.78 | 95.11 |
| AU12 | 81.78 | 88.00 | 90.22 |
| AU15 | 86.67 | 83.11 | 86.67 |
| AU17 | 88.44 | 87.56 | 92.00 |
| AU25 | 94.67 | 92.00 | 93.77 |

AU1 (raised inner eyebrow) scores high classification rate for all experiments. The presence of AU1 is measured by increasing distance between right corner point of left eye contour (P9) and right-most point of upper left eyebrow contour (P27). The distance between two points between male and female facial data has subtle differences. Therefore, all experiments obtained good results.

The distance from left corner of mouth contour (P55) and left face contour (P82) of male and female have significant difference. Thus, the decreased value in male data is less compared to female data. When testing is done using female data, many outliers are detected that eventually reduced the rate of classification.

However, AU12 (mouth corner pulled up) scores lower results across the AUs. AU12 is measured by the distance of P55P82 and P45P55. AU12 is activated if there is decreasing value in both distances. The distance from left corner of mouth contour (P55) and left face contour (P82) of male and female have significant difference. Thus, the decreased value in male data is less compared to female data. When testing is done using female data, many outliers are detected that eventually reduced the rate of classification. Both AU12 and AU15 have lower scores compared to the rest of the AUs. This is due to the fact that distance for the facial points are only significant when the intensity of the expression is high.

Table 5 presents the average classification results using NN and the hidden nodes is set to 10. Cross Entropy (CE) presents the network performance which gives the inaccurate classification. The lower the value of CE, the number of error is reduced. Percent Error (E) specifies the fraction of samples that are classified incorrectly. Zero value indicates no incorrect classification. The neuron is considered good at computing when the CE is value is smaller or near to 0.

TABLE VI. AVERAGE CORRECT CLASSIFICATION USING NN

| AU | Training | | Validation | | Testing | |
|---|---|---|---|---|---|---|
| | CE | E | CE | E | CE | E |
| AU1 | 1.07 | 0.0 | 5.16 | 0.0 | 4.71 | 0.0 |
| AU4 | 3.37 | 0.0 | 9.44 | 0.0 | 9.30 | 0.0 |
| AU6 | 1.06 | 5.16 | 2.71 | 0.0 | 4.95 | 0.0 |
| AU12 | 2.07 | 2.58 | 5.16 | 0.0 | 4.85 | 0.0 |
| AU15 | 1.23 | 2.58 | 3.37 | 0.0 | 3.10 | 0.0 |
| AU17 | 1.59 | 0.0 | 4.76 | 0.0 | 4.78 | 0.0 |
| AU25 | 0.38 | 0.0 | 1.11 | 0.0 | 1.21 | 0.0 |

For AU1, the CE for training (1.31%) is lower compared to validation (5.03%) and testing (4.78%). AU4 obtained lower CE for training compared to validation and testing. For training, only AU25 obtained lowest CE (0.22%) and AU6 scored highest CE (7.65%). For validation, AU25 scored the lowest CE (0.58%) while AU17 scored the highest CE (6.56%). For testing, AU25 obtained the lowest CE (0.56%) while AU17 obtained the highest score (7.23%).

## 6. CONCLUSIONS

In this work, NN and different kernel of SVM are used to classify the recognition rate of AU1, AU4, AU6, AU12, AU15, AU17 and AU25. The performance of SVM is assessed using Linear, Gaussian and Quadratic kernel.

For the future work, the number of features for each AU to be detected will be increased to achieve more accurate result. The distance feature will be measured using different facial points than the present technique used. For example, to detect AU12 (lip corner puller), more distance features could be added such as distance between tip of the chin and left corner of mouth contour. More features provided during the training phase may increase accuracy in AU recognition. Moreover, observation between the relation of increasing the number of data for validation and testing and number of hidden nodes will be conducted.

## ACKNOWLEDGMENTS

This research is fully supported by Malaysian Ministry of Higher Education (MOHE) through Race Acculturation Collaborative Efforts RACE/1331/2016(4). The authors fully acknowledged MOHE and Universiti Malaysia Sarawak for the approved fund which makes this important research viable and effective.